\documentclass[lettersize,journal]{IEEEtran}
\usepackage{amsmath,amsfonts}
\usepackage{algorithmic}
\usepackage{algorithm}
\usepackage{array}
\usepackage[caption=false,font=normalsize,labelfont=sf,textfont=sf]{subfig}
\usepackage{textcomp}
\usepackage{stfloats}
\usepackage{url}
\usepackage{verbatim}
\usepackage{graphicx}
\usepackage{cite}
\usepackage[numbers]{natbib}

\begin{document}

\title{Online Multi-Task Learning with Recursive Least Squares and Recursive Kernel Methods}

\author{Gabriel~R.~Lencione,~\IEEEmembership{Student~Member,~IEEE,} and
        Fernando~J.~Von~Zuben,~\IEEEmembership{Senior~Member,~IEEE}
\thanks{G. R. Lencione and F. J. Von Zuben are with the School of Electrical and Computer Engineering, University of Campinas (Unicamp), Brazil. E-mail:
gabriellencione@gmail.com, vonzuben@unicamp.br}
\thanks{This work has been supported by grants from CNPq - Brazilian National Research Council, proc. \# 308793/2022-6, and Fapesp - The São Paulo Research Foundation, proc. \# 13/07559-3.}}

\markboth{Journal of \LaTeX\ Class Files,~Vol.~14, No.~8, August~2021}%
{Shell \MakeLowercase{\textit{et al.}}: A Sample Article Using IEEEtran.cls for IEEE Journals}


\maketitle

\begin{abstract}
This paper introduces two novel approaches for Online Multi-Task Learning (MTL) Regression Problems. We employ a high performance graph-based MTL formulation and develop two alternative recursive versions based on the Weighted Recursive Least Squares (WRLS) and the Online Sparse Least Squares Support Vector Regression (OSLSSVR) strategies. Adopting task-stacking transformations, we demonstrate the existence of a single matrix incorporating the relationship of multiple tasks and providing structural information to be embodied by the MT-WRLS method in its initialization procedure and by the MT-OSLSSVR in its multi-task kernel function. Contrasting the existing literature, which is mostly based on Online Gradient Descent (OGD) or cubic inexact approaches, we achieve exact and approximate recursions with quadratic per-instance cost on the dimension of the input space (MT-WRLS) or on the size of the dictionary of instances (MT-OSLSSVR). We compare our online MTL methods to other contenders in a real-world wind speed forecasting case study, evidencing the significant gain in performance of both proposed approaches. 
\end{abstract}


\section{Introduction}
\IEEEPARstart{T}{he} success of computer-based inductive learning methods, well represented by the so-called Machine Learning (ML) field, is built on groundbreaking achievements in information technology. With new and more promising techniques, impressive gain in performance and scalability is been testified in a wide range of real-world applications, opening very fertile research fronts.

As a hot topic in ML, Multi-Task Learning (MTL) \cite{survey_mtl} consists in properly modeling and prospecting the relationship of multiple tasks by sharing knowledge during the learning process. The main purpose is to improve the average capability of generalization of the individual tasks. Particularly when the size of the training dataset is reduced, the MTL approach tends to exhibit a distinguished performance in comparison to learning the tasks independently (also known as Single-Task Learning - STL) \cite{saulo}. 

Although different methods have been proposed for modeling the tasks relationship, many of them can be assigned under the so-called Structural Regularization \cite{multi_task_2006}, in which we establish a trade-off between a loss function, which aggregates the individual loss of each task, and a regularization term, which acts as a bias supported by a relational graph. The structural relationship represented in the regularization term is then responsible for promoting information sharing. One of the greatest advantages of this approach is that, once it is associated with linear predictors, most of the final optimization structures are convex, simplifying the training process \cite{multi_task_2006, zhou2011malsar, mtl_2009}. 

Then, let $\mathbf{X}_{t} \in \mathbb{R}^{N_t \times d}$ be the input matrix for task $t \in \{1, 2, …, T\}$, containing $N_t$ training examples of dimension $d$ and let $\mathbf{y}_{t}\in \mathbb{R}^{N_t}$ be the respective target vector. The linear multi-task regression problem is obtained when we attempt to estimate the matrix $\mathbf{\Theta} \in \mathbb{R}^{d\times T}$, whose columns $\mathbf{w}_{t} \in \mathbb{R}^{d}$ represent the individual parameters vectors, in such a manner that the task predictions are given by $\mathbf{\Hat{y}}_{t} = \mathbf{X}_{t}  \mathbf{w}_{t}$. We can therefore state the multi-task learning problem as: 

\begin{equation}
    \mathbf{\Theta}^{*} =  \arg\min_{\mathbf{\Theta}}\mathcal{J}(\mathbf{\Theta}), \mathcal{J}(\mathbf{\Theta}) = \mathcal{L}(\mathbf{\Theta})+\mathcal{R}(\mathbf{\Theta})
\end{equation}
where $\mathcal{L}(\mathbf{\Theta})$ is the common loss function, being most frequently adopted in regression problems as the mean squared error (MSE), which is taken on average over all the $T$ tasks

\begin{equation}
    \mathcal{L}(\mathbf{\Theta}) = \frac{1}{T}\sum_{t=1}^{T}\frac{1}{N_t}\lVert \mathbf{y}_{t} - \mathbf{X}_{t} \mathbf{w}_{t} \rVert_{2}^{2},
\end{equation}
while $\mathcal{R}(\mathbf{\Theta})$ is the regularization term responsible for incorporating tasks relationship during training.

However, when it comes to the applications of ML methods in real-world problems, we can face time-dependent conditions of data that require the models to dynamically adapt over time in order to avoid drastic performance reduction. These conditions may be, for instance, the incremental availability of data to feed the learning algorithms in data streams, often with few or no examples prior to their deployment, or the unanticipated changes in data distribution that may occur in different manners (the so-called Concept Drift and Concept Shift \cite{concept_drift}). Many of the ML (or MTL) pioneer methods were proposed in a batch-fashion learning scheme, in which all the training data are available to be used at once. This type of learning procedure is usually too expensive to be executed at each arrival of a new instance, as for the data stream example, thus motivating the development of recursive methods.

There are, in the literature of Multi-Task Learning, few papers that propose recursive MTL methods, specially for regression problems. As shall be better detailed in the next section, the existing techniques suffer either from slow convergence, resulting in bad approximation to the optimal solution, such as linear algorithms, or from high per-instance cost, sometimes almost as expensive as the batch methods. 

It is even more difficult to find online MTL proposals that implement a nonlinear input-output mapping. The adoption of nonlinearities in ML methods is intended to offer to models more flexibility to fit data that may not be well represented by linear models. The potential benefits of nonlinear Multi-Task Learning have already been outlined in the literature \cite{lencione01}.

This paper comes to fulfill these gaps, with the proposal of two methods for recursive MTL that present better compromise between convergence and computational cost (specially per-instance). We explore the adaptation of two well-known recursive single-task methods to work within a well-succeeded MTL framework. The famous  Weighted Recursive Least Squares (WRLS) and the prominent  Online Sparse Least Squares Support Vector Regression (OSLSSVR) are recursive methods that deal with the same regression problem in the primal and dual spaces, respectively. We demonstrate that a simple modification to the initial conditions of the WRLS, and the appropriate choice for the kernel of the OSLSSVR enable us to achieve two alternative recursive MTL methods with quadratic per-instance complexity and immediate or parameterized convergence. We also combine one of our proposals with Extreme Learning Machines (ELM) \cite{elm}, thus conceiving an online MTL method with nonlinear input-output mapping. We outline our main contributions as follows:
\begin{enumerate}
\item Our paper introduces the Multi-Task Weighted Recursive Least Squares (MT-WRLS), which recursively solves the adopted Graph-based MTL formulation in the primal space, producing the exact solution at each step, thus with immediate convergence, at the per-instance cost of $O(d^2 \times T^2)$.
\item We are the first, to the best of our knowledge, to propose a recursive kernel-based MTL method, the Multi-Task Online and Sparse Least Square Support Vector Regression (MT-OSLSSVR). The approximation to the exact solution can be controlled by its sparsity parameter, presenting a per-instance complexity of $O(d \times m_n
^2)$ ($m_n << n$, $n$ being the number of examples).
\item Both methods are demonstrated to have better theoretical results than most of the existing techniques in literature. With them, we achieve a reasonable per-instance complexity with much stronger guarantees of convergence. 
\item The combination of our MT-WRLS proposal with Extreme Learning Machines enabled the development of a convex online MTL method with nonlinear input-output mapping. 
\item We test our proposals on a real-world time-dependent benchmark of wind speed forecasting and compare them to other contenders in the literature, empirically demonstrating their superior performance.
\end{enumerate}
 
This paper is organized as follows: In Section 2, we present a literature review of the online MTL field. Section 3 is devoted to the formal description of our proposals, followed by their experimental methodology. In Section 4, we present and discuss our results, assessing the achievement of our objectives through a comparative analysis. Sections 5 and 6 provide additional discussions and concluding remarks, also indicating the future steps of our research.

\section{Related Work}

\subsection{Multi-Task Learning via Structural Regularization}

Although there are various proposed multi-task regularizers in the literature, we present three of the most consolidated approaches, motivated by their flexibility and ease of handling:
\begin{itemize}

\item Argryou, Evgeniou \& Pontil \cite{multi_task_2006} proposed to employ the $\ell_{1,2}$ norm of $\mathbf{\Theta}$ as the regularization term, which induces sparsity to the learning model by vanishing entire rows of the parameters matrix. This is sometimes also referred to as group LASSO regularization because of its row-based analogous behaviour. The associated optimization problem becomes:

\begin{equation}
    \mathbf{\Theta} =  \arg\min_{\mathbf{\Theta}}\mathcal{L}(\mathbf{\Theta})+\lambda\lVert \mathbf{\Theta}\rVert_{1,2},
\end{equation}
where $\lambda$ is the hyperparameter controlling the penalty imposed upon the regularization term. 

\item 

Ji \& Ye \cite{mtl_2009_2} supposed that all the tasks share a parameter subspace of reduced dimension. However, since the adoption of $rank(\mathbf{\Theta})$ as the regularization term leads, in general, to an NP-hard problem \cite{semi_2006}, the authors employed the trace norm of the parameter matrix $\lVert \mathbf{\Theta}\rVert_{*}$, that has the property of being an envelope of $rank(\mathbf{\Theta})$. This formulation results in the following optimization problem:

\begin{equation}
    \mathbf{\Theta} =  \arg\min_{\mathbf{\Theta}}\mathcal{L}(\mathbf{\Theta})+\lambda\lVert \mathbf{\Theta}\rVert_{*},
\end{equation}
where $\lambda$ is the respective coefficient, in the same manner as stated in Eq. (3). 
\item
Authors like Ayres and Von Zuben \cite{evep} used a graph-based structure to represent the pairwise tasks relationship. We define the set  $\mathcal{E}$ of edges $\mathbf{e}^{(j)}\in\mathbb{R}^T$, where two tasks (nodes) $x$ and $y$ are related in some degree if $\mathbf{e}_{x}^{(j)}=\alpha$ and $\mathbf{e}_{y}^{(j)}=-\beta$, with $\alpha, \beta > 0$. The whole graph is built upon the matrix $\mathbf{G} = [\mathbf{e}^{(1)}, \ \mathbf{e}^{(2)},\ ...\ , \mathbf{e}^{(\lVert\mathcal{E}\rVert)}]$. The following regularization term penalizes, therefore, the weighted Euclidean distance between every connected pair, opening the possibility of asymmetric connections. 

 \begin{equation}
        \mathcal{R}(\mathbf{\Theta}) = \lVert\mathbf{\Theta G}\rVert_{F}^{2} = \sum_{j=1}^{\lVert\mathcal{E}\rVert}\lVert\mathbf{\Theta }\mathbf{e}^{(j)}\rVert_{2}^{2}
    \end{equation}
  
 The optimization problem becomes:
    
\begin{equation}
        \mathbf{\Theta} =  \arg\min_{\mathbf{\Theta}}\mathcal{L}(\mathbf{\Theta})+\lambda_{1}\lVert\mathbf{\Theta G}\rVert_{F}^{2}+\lambda_{2}\lVert\mathbf{\Theta }\rVert_{2},
    \end{equation}
where the last term corresponds to the norm $\ell_{2}$ of $\mathbf{\Theta}$ and is employed to enforce a general regularization throughout the tasks. The hyperparameters $\lambda_{1}$ and $\lambda_{2}$ control the imposed penalty to the learning process as well. 

In this method, it is important to find efficient ways of determining an appropriate degree of relationship among the tasks, preferably a nonbinary one (e.g. belonging to the [0,1] interval) and also asymmetric, in the sense that the influence of task A on task B can be distinct from the influence of task B on task A. As it has already been done by \cite{evep}, a similarity approach can be pursued, in which the weights $\alpha$ and $\beta$ from a specific edge connecting the tasks $x$ and $y$ are assigned with the similarity measures $sim(x, y)$ and $sim(y, x)$, respectively.
\end{itemize}

Several papers demonstrate how the adoption of MTL can significantly improve the performance of the tasks \cite{survey_mtl}. In the domain of climate variables prediction, we highlight the research of Gonçalves, Von Zuben and Banerjee \cite{goncalves}, who have worked with variables such as temperature, pressure, and humidity. In order to facilitate and disseminate the use of multi-task learning via structural regularization, the MATLAB package MALSAR was proposed \cite{zhou2011malsar}. It implements the three methods previously described (batch formulation) under the aliases of \textit{Least L21}, \textit{Least Trace} and \textit{Least SRMTL}.

\subsection{Online Multi-Task Learning}

We introduce this subsection with a brief discussion about general Online Learning (OL) methods. Many strategies have been developed to efficiently deal with detecting concept drifts and retraining base models in data streams \cite{concept_drift}. Online Learning literature has evolved on a diverse set of branches aiming to face this challenge \cite{online_survey}. A non exhaustive formalism can be stated upon the regret function: let  $\mathcal{J}^{[i]}$ be the loss function and let $\boldsymbol\Theta^{[i]}$ be the learnt parameters of the model at step $i$. The regret at time $n$ expresses the difference between the cumulative losses of the predictions produced by $\boldsymbol\Theta^{[i]}$ and the minimal losses represented by an optimal $\boldsymbol\Theta^*$ considering the same ensemble of losses (all the points until $n$): 

\begin{equation}
    \textit{Regret}(n) = \sum_{i=1}^{n}{\mathcal{J}^{[i]}(\boldsymbol\Theta^{[i]})} - \min_{\boldsymbol\Theta^*}\sum_{i=1}^{n}{\mathcal{J}^{[i]}(\boldsymbol\Theta^*)}
    \label{eq:regret}
\end{equation}

The fundamental idea behind OL is to reduce the average regret ($Regret(n)/n$) over time. When the OL method does not provide an exact solution ($Regret(n)=0$), which could be achieved applying batch learning at each step, the average regret is expected to start at a certain level and to asymptotically converge to zero as the online model learns the general distribution (taken approximately as fixed, at least for a while) from incoming data. The ultimate challenge of Online Learning is to discover the best, if possible, compromise between convergence and computational complexity.

One class of OL algorithms that has gained attention in the last years is the so-called Online Convex Optimization (OCO) \cite{oco_book}. Most of the methods belonging to that class are based on first or second order methods (linear and quadratic cost). The Online Gradient Descent (OGD) is its best exponent since it summarizes the main philosophy of this field: simplicity, online nature, low cost and good convergence rate \cite{oco_book}. It consists of applying a Gradient Descent iteration based on the incoming data instance, thus implementing an online parameters update step. Under the condition of convexity of $\mathcal{J}^{[i]}$, a convergence rate of $O(1/\sqrt{n})$ is guaranteed with a per-instance linear cost.

In \cite{online_2006}, the authors developed an online multi-task perceptron framework for classification problems. Its optimization engine provides, for each task, a combination of the losses of all tasks, resulting in the weights of the parameters update (similar to a learning rate). The structural relationship of the tasks is not totally detailed and, although the update step possesses a linear cost in the dimension of the parameters vector, the cost of finding the update weights depends on the type of the adopted norm. The convergence to an optimal solution is bounded by $O(1/\sqrt{n})$.

Another linear online method was proposed by \cite{online_2010}. It also consists of an online multi-task perceptron for classification problems. The authors proposed a linear update stage in which a matrix $\boldsymbol A$ is employed to incorporate the multi-task relationship into the recursive optimization procedure. It is shown that, according to the choice of matrix $\boldsymbol A$, the number of prediction errors produced by their classifier is bounded by a well-defined structural regularization component. The authors do not specify a convergence rate to the optimal solution. Our proposals are strongly inspired on their approach of dealing with multiple tasks by stacking the parameters vectors into a single vector and on their multi-task kernels.   

The work of \cite{online_2017} is the first, to the best of our knowledge, to start from a batch graph-based multi-task optimization definition and to propose a recursive procedure to approximate the online solution to the optimal one. They presented two different online methods: one profits from a constraint formulation of the optimization problem and applies one iteration of the ADMM (Alternating Direction Method of Multipliers) \cite{DBLP:journals/ftml/BoydPCPE11} algorithm for each online update with a cubic cost on the size of the parameters vectors $O(d^3\times T)$. The other is an implementation of the Online Gradient Descent, presenting a lower cost at the expense of an argued diminished convergence rate, as they also did not provide theoretical convergence rates. The computational cost of the first proposal is similar to exactly solving the batch problem at each iteration. Our research addresses this per-instance burden with exact and approximate recursive solutions.

The authors of \cite{online_2017_2} propose a multi-task framework based on the epsilon tube of the Support Vector Regression (SVR) for modeling ensemble forecasts. Although they describe their method as an online learning procedure, their parameters update step is done in a batch manner with cubic complexity, penalizing the distance between new and current parameters.

More recent works based on OGD and first order algorithms were proposed by \cite{online_2020,online_2020_2}. They explored new structural regularizations and prove theoretical convergence improvements. One important thing to be noticed in all the cited works is either their lack of time-dependent regression case studies, or the absence of fully reproducible and well-described methodologies, making it more difficult to perform a through comparison with the contenders in the literature.

\section{Proposed Methods}

We develop, in this section, our online MTL proposals. First, we detail the reasons behind the adopted MTL formulation, rewriting it to assist in our recursive derivations. Then we present our two recursive methods, one based on the single-task Weighted Recursive Least Squares algorithm and the other based on the Online Sparse Least Squares Support Vector Regression kernel method. 
\subsection{Graph-Based MTL Adoption and Reformulation}
Due to its quadratic optimization structure, convexity, and successful application to synthetic and real-world problems, we selected the graph-based MTL formulation, presented in the last section, for which a recursive version is derived in this work. 

The quadratic shape of the terms composing the function to be minimized, allied to the fact that we are dealing with linear predictors, allows us to convert it to a single task problem with a loss term identical to the least squares linear regression and a quadratic regularization on the parameters vector, making its closed-form solution analogous to the one that is developed for the WRLS, our first base algorithm. 

The existence of a multi-task kernel formalism for the graph-based regularization in the literature \cite{online_2010} also motivates and supports a second online MTL formulation in this work, as we turn to recursive kernel-based methods.

We briefly rewrite the adopted MTL formulation in order to simplify its representation and to promote the understanding of the two alternative recursive proposals.  

Let $\mathbf{X}_\mathit{{t}} \in \mathbb{R}^{n \times d}$ and $\mathbf{y}_\mathit{{t}}\in \mathbb{R}^{n}, t \in 1,...,T$ be the input matrices and output vectors of the $T$ tasks. Let  $\mathbf{\Theta} \in \mathbb{R}^{d\times T} $ be the matrix containing in each column the parameters vector of each task $t$, $\mathbf{w}_\mathit{{t}} \in \mathbb{R}^{d}$, in such a way that the vector of predictions of task $t$ is given by  $\mathbf{\Hat{y}}_{t} = \mathbf{X}_\mathit{{t}} \mathbf{w}_\mathit{{t}}$.

Recalling the general learning process of a multi-task problem, it is usually defined as the minimization of an objective function composed of a loss term, which represents the supervised learning component, and a regularization term, promoting information sharing among tasks \cite{zhou2011malsar}. 

\begin{equation}
    \mathbf{\Theta_{*}} = \arg\min_{\Theta} {J}(\mathbf{\Theta}) 
\end{equation}

\begin{equation}
     J(\mathbf{\Theta}) = \mathcal{L}(\mathbf{\Theta}, \mathbf{X_{1}}, ..., \mathbf{X}_\mathit{{T}}, \mathbf{y_{1}},...,\mathbf{y}_\mathit{{T}}) + \mathcal{R}(\mathbf{\Theta})
\end{equation}

We define, then, the neighbourhood of each task $\mathcal{E}_{t}, t \in 1,...,T$ as the set of tasks that are connected to task $t$ by two-way weighted edges. The weight of the edges is defined by similarities $sim(t,j)$ and $sim(j,t)$ between tasks $t$ and $j$, as done in \cite{evep}, representing the degree of relationship between tasks $t$ and $j$. Therefore, we write the MTL regularization term as: 

\begin{multline}
    \mathcal{R}(\mathbf{\Theta}) = \sum_{t=1}^{T}\sum_{j \in {\mathcal{E}_t}}\lVert \mathbf{w}_\mathit{{t}}sim(t,j) -\mathbf{w}_\mathit{{j}}sim(j,t) \rVert^{2} \\ +\gamma\sum_{t=1}^{T}\lVert\mathbf{w}_\mathit{{t}}\rVert^2\
\end{multline}

Adopting the quadratic error as loss function and the above regularization term, the Graph-Based Multi-Task Learning problem can be stated as follows:
\begin{multline}
    \mathbf{\Theta^{*}} = \arg\min_{\Theta}\sum_{t=1}^{T}\lVert\ \mathbf{X}_\mathit{{t}}\mathbf{w}_\mathit{{t}}- \boldsymbol y_t \rVert^2 \\ +
     \lambda \sum_{t=1}^{T}\sum_{j \in {\mathcal{E}_t}}\lVert\mathbf{w}_\mathit{{t}}sim(j,t)-\mathbf{w}_\mathit{{j}}sim(t,j)\rVert^2 \\+ \gamma\sum_{t=1}^{T}\lVert\mathbf{w}_\mathit{{t}}\rVert^2
     \label{eq:mtl}
\end{multline}

\subsection{Multi-Task Weighted Recursive Least Squares}

In this section, we present our first recursive MTL proposal. As mentioned before, the quadratic shape of the terms composing the MTL objective function motivates the development of a single task closed-form equivalent solution of \eqref{eq:mtl}. We shall see that this closed-form solution is identical to the one that is recursively solved by WRLS, whose iteration demands quadratic cost, presenting therefore, an intermediary complexity with an online exact solution of the problem. 

We start with a brief introduction to the standard Weighted Recursive Least Squares. It will serve as the basic learning structure to be employed in our multi-task framework. 

Let $\mathbf{X}(n) \in R^{n\times d} $ be the input matrix at time $n$, containing $n$ past input vectors of dimension $d$, and let $\mathbf{y}(n) \in R^{n}$ be its respective output vector and $\mathbf{w}(n) \in R^{d}$ the parameters vector at time $n$. The analytical solution of the least squares problem is given by:

\begin{equation}
    \mathbf{w}(n) = \mathbf{\Phi}(n)^{-1}\mathbf{\Psi}(n)
    \label{eq:rls}
\end{equation}
where $\mathbf{\Phi}(n) = (\mathbf{X}(n)'\mathbf{X}(n))$ and $\mathbf{\Psi(n)} = \mathbf{X}(n)'\mathbf{y}(n)$. 

We aim at developing a recursive procedure that solves $\mathbf{\Phi}(n)^{-1}$ and $\mathbf{\Psi}(n)$ from $\mathbf{\Phi}(n-1)^{-1}$ and $\mathbf{\Psi}(n-1)$. As a first step in this direction, we perform the following decomposition with the introduction of the forgetting factor $\sigma$:
\begin{equation}
    \mathbf{\Phi}(n) = \sigma\mathbf{\Phi}(n-1) + \mathbf{x}(n)\mathbf{x}(n)'
\end{equation}
\begin{equation}
        \mathbf{\Psi}(n) = \sigma\mathbf{\Psi}(n-1) + y(n)\mathbf{x}(n)
\end{equation}
where $\mathbf{x}(n) \in R^{d}$ is the input vector at time $n$ and $y(n)$ is the expected output value. Defining  $\mathbf{P}(n) = \mathbf{\Phi}(n)^{-1}$ and employing the Woodbury's identity, we get the following recursive procedure:

\begin{equation}
    \mathbf{k}(n) = \frac{\mathbf{P}(n-1)\mathbf{x}(n)}{\sigma+\mathbf{x}(n)'\mathbf{P}(n-1)\mathbf{x}(n)}
\end{equation}
\begin{equation}
    \alpha(n) = y(n) - \mathbf{x}(n)'\mathbf{w}(n-1)
\end{equation}
\begin{equation}
    \mathbf{w}(n) = \mathbf{w}(n-1) +  \alpha(n)\mathbf{k}(n)
\end{equation}
\begin{equation}
    \mathbf{P}(n) = \sigma^{-1}[\mathbf{P}(n-1) - \mathbf{k}(n)\mathbf{x}(n)'\mathbf{P}(n-1)]
\end{equation}

The last step to complete the WRLS algorithm is to choose the initial conditions $\mathbf{w}(0)$ and $\mathbf{P}(0)$, which are usually defined as:
\begin{itemize}
    \item $\mathbf{w}(0) = \mathbf{0}$ 
    \item $\mathbf{P}(0) = \lambda \textbf{I}_{d\times d}$, $\lambda>0$, to guarantee non-singularity of the initial $\mathbf{P}(n)$.
\end{itemize}

Recalling our graph-based MTL formulation \eqref{eq:mtl}, we derive, in the next paragraphs, a closed-form solution for this optimization problem in order to develop a multi-task WRLS. Since it is a convex unrestricted optimization problem \cite{zhou2011malsar}, we shall only consider the third KKT condition, which guarantees that the stationary points of its Lagrangian are global minima. 

We seek, therefore, the parameters for which the gradient of the objective function is zero.  

\begin{multline}
    \nabla_{\mathbf{w}_\mathit{{t}}} J(\mathbf{\Theta}) = -2\mathbf{X}_\mathit{{t}}'\mathbf{y}_\mathit{{t}}+2\mathbf{X}_\mathit{{t}}'\mathbf{X}_\mathit{{t}}\mathbf{w}_\mathit{{t}}\\ + 2\lambda\sum_{j \in {\mathcal{E}_t}}[\mathbf{w}_\mathit{{t}}sim(t,j)-\mathbf{w}_\mathit{{j}}sim(j,t)] \\ + 2\lambda\gamma\mathbf{w}_\mathit{{t}}, t \in 1,...,T
\end{multline}
\begin{multline}  
    \nabla_{\mathbf{w}_\mathit{{t}}} J(\mathbf{\Theta}) = \mathbf{0}, t \in 1,...,T
    \\ \implies   \mathbf{X}_\mathit{{t}}'\mathbf{X}_\mathit{{t}}\mathbf{w}_\mathit{{t}}+ \lambda\sum_{j \in {\mathcal{E}_t}}[\mathbf{w}_\mathit{{t}}sim(t,j)-\mathbf{w}_\mathit{{j}}sim(j,t)] \\+ \lambda\gamma\mathbf{w}_\mathit{{t}} = \mathbf{X}_\mathit{{t}}'\mathbf{y}_\mathit{{t}}
    \\ \implies  
    \{\mathbf{X}_\mathit{{t}}'\mathbf{X}_\mathit{{t}} + \lambda[\gamma+\sum_{j \in {\mathcal{E}_t}}sim(t,j)]\mathbf{I}\}\mathbf{w}_\mathit{{t}}\\ -\lambda\sum_{j \in {\mathcal{E}_t}}\mathbf{w}_\mathit{{j}}sim(j,t) = \mathbf{X}_\mathit{{t}}'\mathbf{y}_\mathit{{t}}
    \label{eq:statio}
\end{multline}

We now perform a transformation to assist in developing our solution. Instead of working with a matrix $\mathbf{\Theta}$ and different input matrices for each task, we stack all the parameters vectors of each task into a single stacked vector, all the output vectors into a single output vector and all the input matrices into a block-diagonal input matrix, defined as follows:

\begin{multline}
\mathbf{X} = 
    \begin{bmatrix}
    \mathbf{X}_{{1}}&\mathbf{0}&....&\mathbf{0}\\
    \mathbf{0}&\mathbf{X}_{{2}}&....&\mathbf{0}\\
    .&.&.&.\\
    .&.&.&.\\
    .&.&.&.\\
    \mathbf{0}&\mathbf{0}&....&\mathbf{X}_\mathit{{t}}
    \end{bmatrix}
     \in R^{nT\times dT}, \\ \mathbf{y} = 
     \begin{bmatrix} 
    \mathbf{y}_{{1}}\\
    \mathbf{y}_{{2}}\\
    .\\.\\.\\
    \mathbf{y}_{{T}} 
    \end{bmatrix}
    \in R^{nT}, 
    \mathbf{w} = \begin{bmatrix} 
    \mathbf{w}_{{1}}\\
    \mathbf{w}_{{2}}\\
    .\\.\\.\\
    \mathbf{w}_{{T}}
    \end{bmatrix} 
    \in R^{dT}
\end{multline}

 Adopting this new representation scheme of our MTL basic elements, it is trivial to see that the set of equations in Eq. \eqref{eq:statio} can be rewritten as: 

\begin{equation}
    \{\mathbf{X'}\mathbf{X} + \lambda\boldsymbol A\otimes \boldsymbol I_{d} \}\mathbf{w}  = \mathbf{X}'\mathbf{y}
\end{equation}
where $\boldsymbol A\otimes \boldsymbol I_{d}$ is the Kronecker product of matrix $\boldsymbol A \in R^{T\times T}$ (defined below) and the identity matrix.
\begin{equation}
    \tiny
    \boldsymbol A =
    \begin{bmatrix}
 \gamma+\sum_{j \in {\mathcal{E}_1}}sim(1,j)&-sim(1,2)&....&-sim(1,T)\\
    -sim(2,1)&\gamma+\sum_{j \in {\mathcal{E}_2}}sim(2,j)&....&-sim(2,T)\\
    .&.&.&.\\
    .&.&.&.\\
    .&.&.&.\\
    -sim(T,1)&-sim(T,2)&....&\gamma+\sum_{j \in {\mathcal{E}_T}}sim(T,j)
    \end{bmatrix}
    \label{eq:a}
\end{equation}

Thus, the closed-form solution for the MTL parameters vector becomes: 
\begin{equation}
     \mathbf{w}  = \{\mathbf{X'}\mathbf{X} + \lambda\boldsymbol A\otimes \boldsymbol I_{d} \}^{-1}\mathbf{X}'\mathbf{y}
\end{equation}
which has the same form of \eqref{eq:rls}. If we state that $\mathbf{P}(0)=\{\lambda\boldsymbol A\otimes \boldsymbol I_{d}\}^{-1}$, then we can make every new task input vector as an input vector of dimension $dT$ with zeros in all indices other than those of the specific task. The stacked input vectors are then inputs to the WRLS algorithm.  

The computational complexity of WRLS at each iteration for each task is $O(d^2)$. When we operate on the stacked multi-task space, this complexity becomes $O(d^2\times T^2)$, which is lower than most nonlinear approximated solutions presented in literature. Its initialization requires the computation of $\{\lambda\boldsymbol A\otimes \boldsymbol I_{d}\}^{-1} = \{\lambda^{-1}\boldsymbol A^{-1}\otimes \boldsymbol I_{d}\}$ whose cost is usually $O(T^3)$. It may seem as expensive as the other proposals. However, this cubic burden is, on the other hand, only required once before the arrival of the first instances into the MT-WRLS algorithm.

\begin{algorithm}[H]
\caption{Multi-task Weighted Recursive Least Squares}\label{alg:MT-WRLS}
\begin{algorithmic}
\STATE 
\STATE {Define the similarities} $sim(t,s); t,s \in 1,2,...,T$
\STATE {Set} $\gamma, \lambda > 0$ and $i\gets0$
\STATE {Compute $\boldsymbol{A}$ 
 according to Eq. \eqref{eq:a}} 
 \STATE {Set $\textbf{P}(0) \gets \lambda^{-1}\boldsymbol A^{-1}\otimes \boldsymbol I_{d}$ } 
 \STATE {Set $\textbf{w}(0) \gets \textbf{0}_{dT} $ } 
\STATE $ \textbf{for}$ $n=1,2, ...$ $ \textbf{do} $
\STATE \hspace{0.5cm}$ \textbf{for}$ $t=1,2, ..., T$ $ \textbf{do} $
\STATE \hspace{1cm} Set $i \gets i + 1$
\STATE \hspace{1cm} Predict $\hat{y}_t(n)$ with $\textbf{w}_t(i-1)$ and $\textbf{x}_t(n)$
\STATE \hspace{1cm} Set $ \textbf{x}(i)$ as the stacked vector of $\textbf{x}_t(n)$
\STATE \hspace{1cm} Compute $\textbf{k}(i)$ according to Eq. (15)
\STATE \hspace{1cm} Compute $\boldsymbol{\alpha}(i), \textbf{w}(i)$ using Eqs. (16) and (17)
\STATE \hspace{1cm} Compute $\boldsymbol{P}(i)$ resorting to Eq. (18)
\end{algorithmic}
\label{alg1}
\end{algorithm}

\subsection{Multi-Task Online Sparse Least Squares Support Vector Regression}

Kernel methods have established their fame and success due mostly to their ability to map, linearly or nonlinearly, the input space onto a feature hyperspace of finite or infinite dimension. The kernel function, computed directly on the original input space, can provide enough information to allow the methods to learn very difficult patterns. Since they usually work on the dual space, their batch-style optimization complexity is commonly defined by $O(n^3)$ ($n$ being the number of training examples), which does not scale well with the increasing amount of instances provided by data streams. 

Recursive kernel methods for regression were proposed to easy this burden, but none of them have yet considered an online multi-task approach. As done in the MT-WRLS case, just proposed in the previous section, we are going to select a method that recursively solves a general single task static problem, followed by its adaptation to our MTL framework. In this case, it is done by choosing the appropriate kernel function and demonstrating the equivalence between the adapted optimization problem and our graph-based MTL original formulation. 

First, let us define the Least Squares Support Vector Regression problem as done in \cite{oslssvr}. It is based on the dual formulation of the least squares problem and considers all of the examples in the training set as support vectors. Its general optimization structure is given by:

\begin{equation}
    \mathbf{\boldsymbol\alpha_{*}} =  \arg\min_{\boldsymbol\alpha}\lVert\mathbf{K\boldsymbol\alpha-y}\rVert^{2}+\lambda\mathbf{\boldsymbol\alpha' K \boldsymbol\alpha}
    \label{eq:oslssv}
\end{equation}
\begin{equation}
    \textbf{ŷ} = \textbf{K}\boldsymbol\alpha
\end{equation}
where $\boldsymbol\alpha \in \mathbb{R}^{n}$ is the parameters vector, $n$ the number of examples, and $\mathbf{K} \in \mathbb{R}^{n\times n}$ is the kernel matrix defined as follows:

\begin{equation}
    \mathbf{K}_{[i,j]} = \langle\boldsymbol\phi(\mathbf{x}_i),\boldsymbol\phi(\mathbf{x}_j)\rangle
\end{equation}
with $\boldsymbol\phi(\mathbf{x}_i)$ being the mapping of  $\mathbf{x}_i$ ($i$-th row of $\mathbf{X}$) onto an specific hyperspace. 

The solution to this optimization problem is given by its stationary point:
\begin{equation}
\boldsymbol\alpha_{*}=(\mathbf{K}+\lambda \mathbf{I})^{-1} \mathbf{y},
\label{eq:6}
\end{equation}
whose computation requires solving a linear system on the number of examples. 

The work of \cite{oslssvr} proposes a recursive solution for $\mathbf{\boldsymbol\alpha_{*}}(n)$ (see Eq. \eqref{eq:oslssv}) (Online Sparse Least Squares Support Vector Regression -  OSLSSVR), having $\textbf{K}(n)$ and $\textbf{y}(n)$ as time-dependent elements of the problem. Their contribution is incremental in the sense that a recursive solution of Eq. \eqref{eq:oslssv} with $\lambda=0$ was already proposed by \cite{krls} with the so-called Kernel Recursive Least Squares (KRLS) method. In fact, $\lambda>0$ includes a regularization term into the learning procedure, which is  fundamental for us to achieve a dual equivalence of Eq. \eqref{eq:mtl} and, therefore, apply the OSLSSVR method to obtain a recursive kernel-based multi-task learning method. This equivalence is supported by the Representer Theorem and is detailed in the next paragraphs.

The OSLSSVR method was designed to solve the optimization problem in Eq. \eqref{eq:oslssv} in an online manner, thus avoiding the storage of all data instances since the linear system in Eq. \eqref{eq:6} turns out to be solved for every new incoming instance. It employs a dictionary containing only a subset of the input
instances $\mathcal{D}_{n}^{s v}=\left\{\tilde{\mathbf{x}}(m)\right\}_{m=1}^{m_{n}}$, capable of spanning the feature hyperspace within some margin of error, represented by the Approximate Linear Dependency (ALD) criterion \cite{krls}, which imposes the condition of inserting a new training input to the dictionary in the following way 

\begin{equation}
 \delta_{n} \stackrel{\operatorname{def}}{=} \min _{\mathbf{a}}\left\|\sum_{m=1}^{m_{n-1}} a_{m} \mathbf{\phi}\left(\tilde{\mathbf{x}}(m)\right)-\mathbf{\phi}\left(\mathbf{x}(n)\right)\right\|^{2} \leq v
 \label{eq:7}
\end{equation}
where $v$ is the sparsity level parameter and $\mathbf{a}=\left(a_{1}, \ldots, a_{m_{n-1}}\right)^{T}$ is the vector containing the coefficients of the linear combination. We state that the input instance at time $n$ can be approximated by a linear combination of the current dictionary within a squared error $v$ if $\delta_{n} \leq v$. Hence, the kernel matrix at step $n$ can be computed using only the elements of $\mathcal{D}_{n}$. 

Once this approximated kernel matrix is adopted, the OSLSSVR method provides the exact recursive solutions to its learning problem with per-instance cost of $O(m_n^2)$, making the convergence to the optimal solution be regulated by $v$. With greater values of $v$, we add less input samples to the dictionary and the kernel estimation may not be highly accurate. On the other hand, a $v$ tending to zero promotes the dictionary to tend to the whole set of examples, augmenting the computational cost $O(m_n^2) \rightarrow O(n^2)$.  

Let $\mathbf{W} \in R^{dT}$ be the stacked parameters vector of the $T$ tasks and let $\mathbf{x}_{{(n,s)}}=(0,...,\mathbf{x}_{s}{(n)}',...,0)' \in R^{dT}$ be the stacked input vector of task $s$ at time $n$. Let still the RKHS $H$ of dimension $dT$ with inner product $\langle\boldsymbol u,\boldsymbol v\rangle_H = \boldsymbol u'(\boldsymbol A\otimes \boldsymbol I_{d}) \boldsymbol v$, with $\boldsymbol A$ symmetric and positive-definite. The mapping function of the stacked input vector onto the RKHS is defined by:
\begin{equation}
     \boldsymbol \phi: R^{dT} \rightarrow H, \boldsymbol \phi(\mathbf{x}_{{(n,s)}}) = \boldsymbol A\otimes^{-1}\mathbf{x}_{{(n,s)}}
\end{equation}
for simplicity, we write $\boldsymbol A\otimes \boldsymbol I_{d}$ as $\boldsymbol A\otimes$ and $\boldsymbol A^{-1}\otimes \boldsymbol I_{d}$ as $\boldsymbol A\otimes^{-1}$

The kernel function of two input vectors of any tasks $s$ and $t$, at any time points $n$ and $l$, is given by:
\begin{multline}
    K(\mathbf{x}_{s}{(n)}, \mathbf{x}_{t}{(l)}) = \langle\phi(\mathbf{x}_{{(n,s)}}), \phi(\mathbf{x}_{{(l,t)}})\rangle \\= (\mathbf{x}_{{(n,s)}}'\boldsymbol A\otimes^{-1}) \boldsymbol A\otimes (\boldsymbol A\otimes^{-1}\mathbf{x}_{{(l,t)}}) \\
=\mathbf{x}_{{(n,s)}}'\boldsymbol A\otimes^{-1}\mathbf{x}_{{(l,t)} }= \mathbf{x}_{s}{(n)}'\mathbf{x}_{t}{(l)}
    \boldsymbol A ^{-1} _{[s,t]}
    \label{eq:kernel}
\end{multline}

We can see that the kernel function can be easily computed by the inner product of the input vectors in their original space ($dim=d$) multiplied by a factor that depends on their relationship. It will allow us to avoid computations on $R^{dT}$, implying in cost reduction. 

Let us demonstrate that, with the appropriate choice for $A$, we can turn the multi-task optimization problem into the same kernel formulation as presented in Eq. \eqref{eq:oslssv}, enabling us to employ the OSLSSVR algorithm as a recursive multi-task method.  

The Representer Theorem guarantees  that, under certain conditions, there is a kernel-based solution to the Empirical Risk Minimization problem:   

\begin{equation}
    f_{*} =  \arg\min_{f}\lVert f(\mathbf{X})-\mathbf{y}\rVert^2+\lambda\lVert f\rVert_{H}^{2}
    \label{eq:repre1}
\end{equation}

\begin{equation}
    f_{*} = \sum_{t=1}^{T}\sum_{j}^{n} \alpha_{t,j} K(\mathbf{x}_{t}{(j)}, .)
    \label{eq:repre2}
\end{equation}
where we explicitly wrote double summations only to represent the entire dataset of all tasks stacked. 

It is well known \cite{oslssvr} that the combination of Eqs. \eqref{eq:repre1} and \eqref{eq:repre2} results in Eq. \eqref{eq:oslssv}, for which the parameters to be optimized are those associated with the support vectors. We now show that Eq. \eqref{eq:repre1} is equivalent to our MTL formulation in Eq. \eqref{eq:mtl}. 

Writing $f$ in terms of a parameters vector $\boldsymbol E$ in the hyperspace:
\begin{multline}
 f = \sum_{t=1}^{T}\sum_{j}^{n} \alpha_{t,j} K(\mathbf{x}_{t}{(j)} , .) \\= \sum_{t=1}^{T}\sum_{j}^{n} \alpha_{t,j} \langle\boldsymbol\phi(\mathbf{x}_{{(j,t)}}) , \boldsymbol \phi(.)\rangle  \\= \langle \sum_{t=1}^{T}\sum_{j}^{n} \alpha_{t,j}\boldsymbol \phi(\mathbf{x}_{{(j,t)}}) , \boldsymbol \phi(.)\rangle \\= \langle \boldsymbol e , \boldsymbol \phi(.)\rangle
\end{multline}
 However $\langle\boldsymbol e, \boldsymbol \phi(\mathbf{x}_{{(n,s)}})\rangle =  \langle\boldsymbol e ,\boldsymbol A\otimes^{-1} \mathbf{x}_{{(n,s)}}\rangle=\boldsymbol e' \boldsymbol A\otimes \boldsymbol A\otimes^{-1} \mathbf{x}_{{(n,s)}} = \boldsymbol e' \mathbf{x}_{{(n,s)}}$. Hence, the parameters vector that acts in the hyperspace is equivalent to the vector $\mathbf{w}$ and, therefore, $f(\mathbf{X}) = \boldsymbol X\mathbf{w}$.

If we adopt $\boldsymbol A$ as done before in the MT-WRLS section (see Eq. \eqref{eq:a}), then we obtain:
\begin{multline}
    \lVert f\rVert_{H}^{2} = \lVert \boldsymbol e\rVert^{2} = \langle\boldsymbol e, \boldsymbol e\rangle = \boldsymbol e' \boldsymbol A\otimes \boldsymbol e \\ =  \sum_{t=1}^{T}\sum_{j \in {\mathcal{e}_t}}\lVert\mathbf{w}_\mathit{{t}}sim(j,t)-\mathbf{w}_\mathit{{j}}sim(t,j)\rVert^2 \\ + \gamma\sum_{t=1}^{T}\lVert\mathbf{w}_\mathit{{t}}\rVert^2
\end{multline}
    
Therefore, Eq. \eqref{eq:repre1} can be rewritten as:
\begin{multline}   
     f_* = \arg\min_{f}\lVert\ \boldsymbol X\mathbf{w} - \boldsymbol Y \rVert^2 \\ +
     \lambda\{ \sum_{t=1}^{T}\sum_{j \in {\mathcal{e}_t}}\lVert\mathbf{w}_\mathit{{t}}sim(j,t)-\mathbf{w}_\mathit{{j}}sim(t,j)\rVert^2 \\ + \gamma\sum_{t=1}^{T}\lVert\mathbf{w}_\mathit{{t}}\rVert^2\}
\end{multline}
which is fully analogous to our graph-based MTL formulation. 

Then, we found a multi-task kernel (defined by $\boldsymbol A$) which can be adopted in the OSLSSVR algorithm to recursively learn and promote sharing information according to the degree of relationship among the tasks. As the iteration cost of OSLSSVR method is $O(m_{n}^{2})$, with $m_{n}$ being the size of the support vectors dictionary (usually $m_{n}<< nT$ or limited to a constant), 
multiplied by the complexity of computing the kernel function, which is $O(d)$, then the per-instance cost of MT-OSLSSVR is $O(d \times m_{n}^{2})$.   

The major advantage of this method in comparison to the proposed MT-WRLS is that the multi-task kernel can be computed using individual task input vectors of dimension $d$, instead of working on the stacked space of dimension $dT$.

\begin{algorithm}[H]
\caption{Multi-Task Online Sparse Least Squares Support Vector Regression}\label{alg:MT-OSLSSVR}
\begin{algorithmic}
\STATE 
\STATE {Define the similarities} $sim(t,s); t,s \in 1,2,...,T$
\STATE {Set} $\gamma, \lambda > 0$ and $i\gets0$
\STATE {Compute $\boldsymbol{A}$ 
 according to Eq. \eqref{eq:a}} 
  \STATE {Set the kernel function using Eq. \eqref{eq:kernel}} 
 \STATE {Initialize the OSLSSVR algorithm according to \cite{Lencione2023}} 
 
\STATE $ \textbf{for}$ $n=1,2, ...$ $ \textbf{do} $
\STATE \hspace{0.5cm}$ \textbf{for}$ $t=1,2, ..., T$ $ \textbf{do} $
\STATE \hspace{1cm} Predict $\hat{y}_t(n)$ with $\textbf{x}_t(n)$ and OSLSSVR \cite{Lencione2023}
\STATE \hspace{1cm} Train OSLSSVR \cite{Lencione2023} with $\textbf{x}_t(n)$ and ${y}_t(n)$
\end{algorithmic}
\label{alg1}
\end{algorithm}

\section{Experimental Setup}

\subsection{Online MTL Contenders}

We implement two online MTL contenders in order to be able to partially rank our proposals among the literature of online Multi-Task Learning, which was demonstrated to be lacking of reproducible time-dependent regression experiments. These methods are the Online Gradient Descent of our graph-based formulation, which was already proposed by \cite{lencione02} (hereinafter referred to as MOGD), and the proposal of \cite{online_2017} presented before as an online single-run of the ADMM optimization procedure (MADMM).

\subsection{Nonlinear Online MTL with Extreme Learning Machines}
We also combine the original WRLS and our MT-WRLS proposal with Extreme Learning Machines (ELM), which are Single Hidden Layer Feedforward Neural Networks (SHLFNs) \cite{elm} capable of providing a nonlinear input-output mapping with a simpler training process, sampling the hidden layer weights and estimating the output layer weights via ordinary least squares or some sort of regularized regression. 

Let $\mathbf{x}_\mathit{{t}}\in \mathbb{R}^d$ be an input vector from task $t$. The respective SHLFNs output prediction is given by:
\begin{equation}
    \hat{y_t} = \sum_{k=1}^{H}\beta_{t,k} f(\mathbf{x}_\mathit{{t}}^{T} \mathbf{v}_{k})+\beta_{t,0}, 
\end{equation}
where $H$ is the number of hidden nodes, $\mathbf{v}_{k} \in \mathbb{R}^d$ is the hidden layer weight vector of node $k$, $f$ is the nonlinear activation function and $\boldsymbol{\beta_t}=[\beta_{t,0},\beta_{t,1}, \beta_{t,2},..., \beta_{t,H}] \in \mathbb{R}^{H+1}$ is the parameters vector of the output layer.

The number of neurons in the hidden layer, the sampling method of their weights and the adoption of the hyperbolic tangent as activation function are important parameters of the ELM that may have a direct impact on its performance. We follow the same methodology to deal with these aspects as done by \cite{lencione01}.

\subsection{Online Regression Benchmark}
Due to the lack of multi-task time-dependent regression benchmarks, we only test our proposals on the wind speed forecasting dataset proposed by \cite{lencione01}. This case study consists of $T=10$ time series of wind speed from wind sites located in Miami, United States. They were extracted from the Wind Integration National Dataset (WIND) Toolkit \cite{DRAXL2015355}, which is an important meteorological database of thousands of wind sites from all over the U.S. territory, with data collected every 5 minutes from 2007 up to 2012. The wind speed intensities were registered at the height of 100 meters and are expressed in meters per second (m/s). Aiming to perform one-step-ahead forecasts (5 minutes), the authors adopted the first three months of 2012, resulting in approximately 24 thousand points for each time series. 

We pursued the proposed methodology and sampled 30 multi-task subsets $C_{k}$ ($k \in {1,2...,30}$) from the 10 original time series, ending up with $30$ multi-task datasets of wind speed prediction containing $10$ tasks of $400$ points each, which allows us to investigate the accuracy of the proposed methods in $30$ different benchmarks. 

\begin{figure*}[!t]
\centering
\subfloat[]{\includegraphics[width=3.5in]{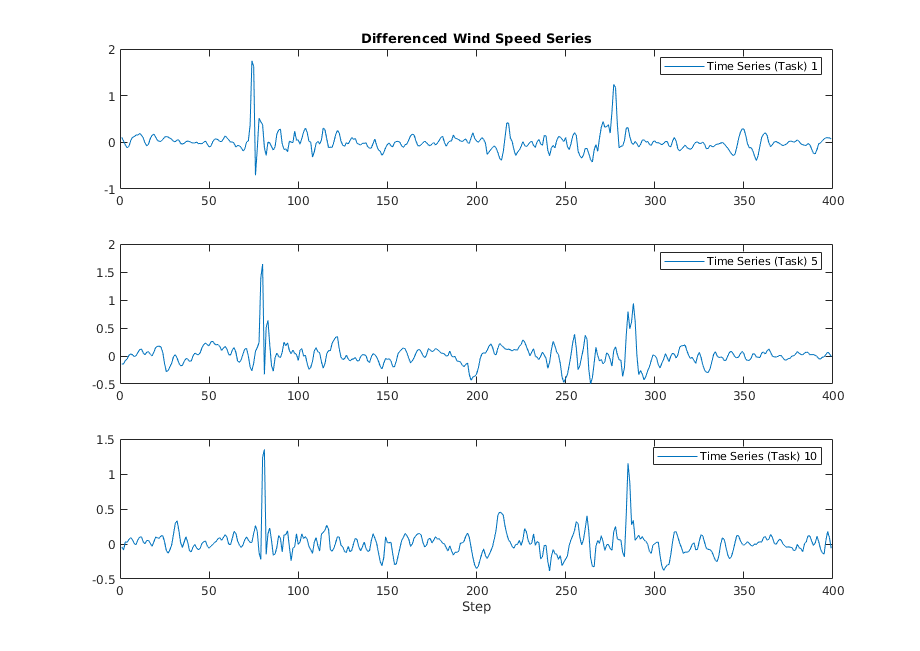}%
\label{fig_first_case}}
\hfil
\subfloat[]{\includegraphics[width=3.5in]{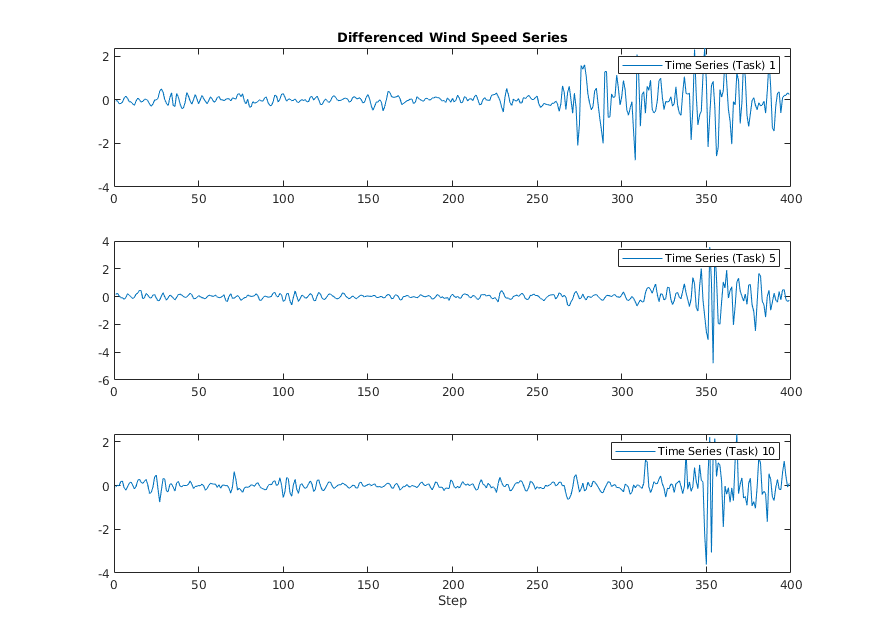}%
\label{fig_second_case}}
\caption{Differentiated wind speed series 1, 5 and 10, each one representing a prediction task $t$, for subsets (a) $C_{1}$ (b) $C_{13}$.}
\label{fig1}
\end{figure*}

\begin{figure*}[!t]
\centering
\subfloat[]{\includegraphics[width=3.5in]{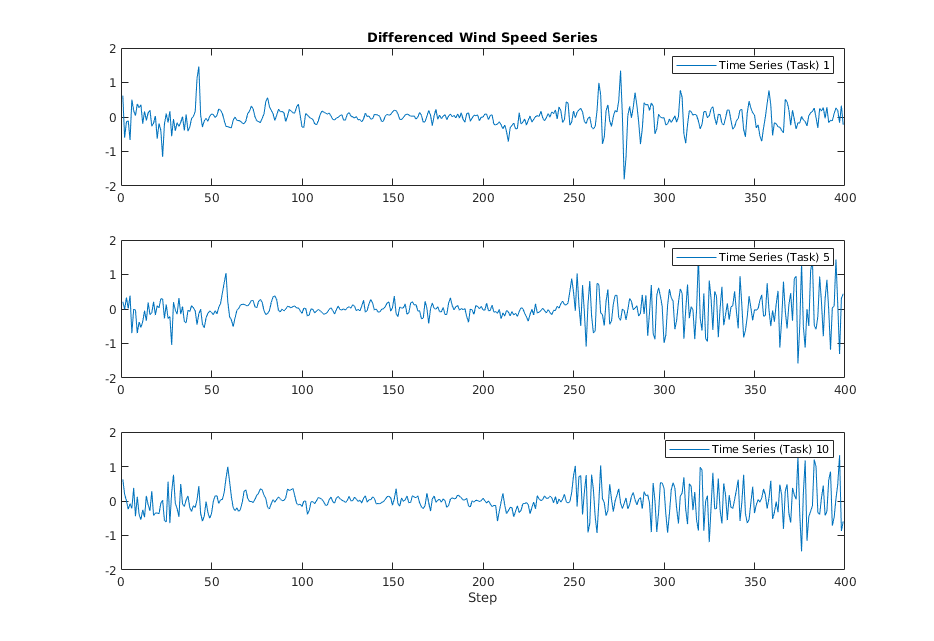}%
\label{fig_first_case}}
\hfil
\subfloat[]{\includegraphics[width=3.5in]{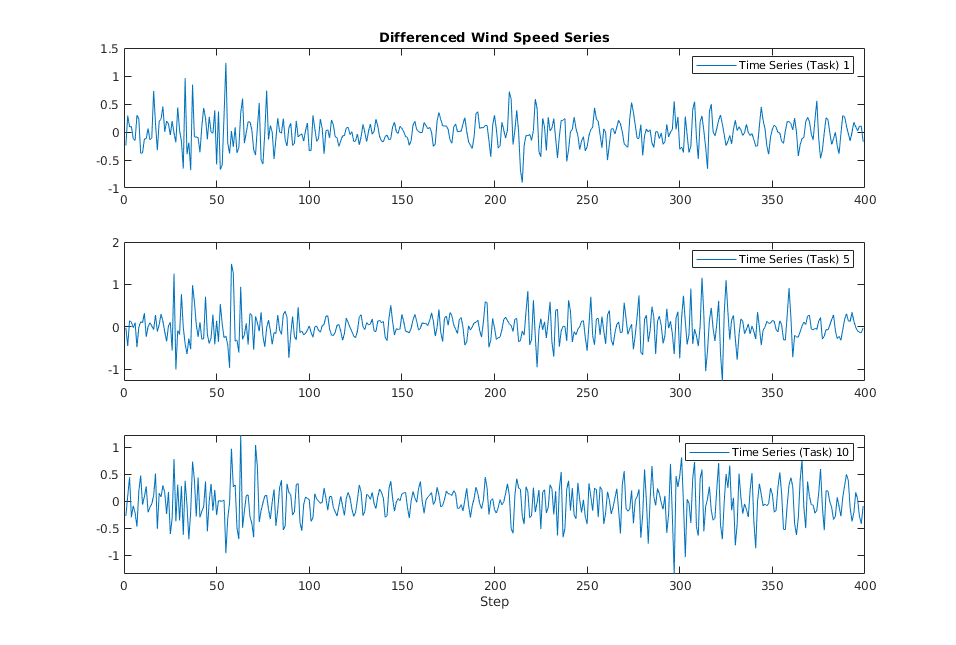}%
\label{fig_second_case}}
\caption{Differentiated wind speed series 1, 5 and 10, each one representing a prediction task $t$, for subsets (a) $C_{23}$ (b) $C_{29}$.}
\label{fig2}
\end{figure*}

Fig. \ref{fig1} and \ref{fig2} illustrate three differentiated wind speed series, $1$, $5$ and $10$ (each one seen as a prediction task) for four different subsets, $C_1, C_{13}, C_{23}$ and $C_{29}$. As we may note, the time series present a variety of patterns along time, reinforcing the need for efficient online learning methods capable of handling changes in data distribution. On the other hand, for each subset $C_k$, the similarity between the time series becomes evident, encouraging even more the adoption of multi-task learning. 

The authors also aimed at studying the ability of MTL methods to better generalize in the presence of less training data as mentioned before, adopting, for each subset $C_{k}$, two different percentages of sequential training and test split: $\mu \in \{27.5\%, 45\%\}$. We shall refer to the first percentage as Experiment I and the second one as Experiment II.

In their work \cite{lencione01}, the adoption of first-order differentiation to remove linear trends was responsible for achieving higher general performance, which motivated us to only work with differentiated series in this experiment. To ease the burden of selecting the best size of the auto-regressive vector, we set the past $9$ occurrences of the series as the input for the step-ahead predictor as follows:
\begin{multline}
   \Hat{y}^{[i]} = w_1y^{[i-1]} + w_2y^{[i-2]} + w_3y^{[i-2]} +  w_4y^{[i-4]}  + w_5y^{[i-5]} \\ + w_6y^{[i-6]} + w_7y^{[i-7]} + w_8y^{[i-8]} + w_9y^{[i-9]} + w_0.
\end{multline}
where the max lag for the auto-regressive vector is set to $9$, as our preliminary experiments, following the same methodology of \cite{lencione01}, indicated to be the most frequent result. This is done in order to reduce the number of hyperparameters to be tuned, improving the parsimony and the understanding of our experimental setup.


\subsection{Optimization of Hyperparameters}

For each subset $C_{k}$ and train/test split percentage $\mu$, we optimize the hyperparameters of the recursive methods on training set and run the online method on test set, as done in \cite{lencione01}. Then, the training set is not used to initialize the parameters of the online methods, which shall be learnt from scratch on test set. The pairwise similarities were computed by the Spearman Correlation between the series in the training sets.

For the MT-WRLS proposal, we tune the hyperparameters $\sigma$ (forgetting factor) and $\lambda$ (penalization of the regularization term) within the ranges $\{0.01, 0.2, 0.4, 0.6, 0.8, 1.0\}$ and $\{10^{-10}, 10^{-4},10^{-3},10^{-2},10^{-1}$, $10^{0},10^{1},10^{2},10^{3},10^{4},10^{10}\}$, respectively. For MT-OSLSSVR method, we tune $\nu$ (sparsity control of OSLSSVR \cite{oslssvr}) and $\lambda$ (penalization of the regularization term) within $\{10^{-3},10^{-2},10^{-1}\}$ and $\{10^{-10}, 10^{-4},10^{-3},10^{-2},10^{-1}$, $10^{0},10^{1},10^{2},10^{3},10^{4},10^{10}\}$, respectively.

\subsection{Procedures for Comparing our Proposals with Contenders}

We compare our proposals with online methods, original single-task WRLS and OSLSSVR, and with batch methods, those studied in \cite{lencione01}. The relative root mean square error (RELRMSE) and the relative mean absolute error (RELMAE), which are not scale sensitive, are the metrics employed by the authors to summarize the performance of all $T$ tasks in each $C_{k}$, avoiding comparison in different scales. They are computed as the ratio of the root mean square error (RMSE) and the mean absolute error (MAE) produced by the predictions of the experimented methods and those produced by a benchmark model, detailed below.

\begin{equation}
   RELRMSE(t,k,\mu) = \frac{RMSE(t,k,\mu)}{RMSE_{persistence}(t,k,\mu)}
\end{equation}

\begin{equation}
   RELMAE(t,k,\mu) = \frac{MAE(t,k,\mu)}{MAE_{persistence}(t,k,\mu)}
\end{equation}
where the adopted benchmark is the persistence method, whose one-step-ahead forecasts are given simply by $\Hat{y}_{t}^{[i]} = y_{t}^{[i-1]}$.

\begin{equation}
   RMSE(t,k,\mu) = \sqrt{\frac{1}{n_{\mu}}\sum_{i=1}^{n_{\mu}}({\Hat{y}_{t}^{[i]}}-y_{t}^{[i]})^{2}}
\end{equation}

\begin{equation}
   MAE(t,k,\mu) = \frac{1}{n_{\mu}}\sum_{i=1}^{n_{\mu}}\lvert{\Hat{y}_{t}^{[i]}}-y_{t}^{[i]}\rvert
\end{equation}
where $n_{\mu}$ is the length of the respective test set. 

We compute the mean metrics over all the $T$ tasks in order to assess the average performance of each method per experiment (training percentage). Once we have the results for our two alternative proposals and the other contenders, we submit the mean $RELRMSE(k,\mu)$ to the Friedman Test followed by the post-hoc Fisher Test (or LSD test) \cite{Pereira_Afonso_Medeiros_2015}, which are non-parametric hypothesis tests designed to measure the statistical significance of the difference between the ranks of distinguished treatments. Having different mean ranks at a significance level of $0.05$, we account a victory to the learning method with a lower mean rank and a defeat for the other.

\section{Results and Discussion}
We present, in Table \ref{tab:table1}, the mean RELRMSE and mean RELMAE for each contender of Experiment I. As we shall see, the MT-WRLS, MT-WRLS+ELM and MT-OSLSSVR were the ones with the lowest errors among the investigated methods, providing up to $16\%$ of accuracy improvement when comparing to their STL versions. Our proposals also surpass the performance of other multi-task recursive algorithms, demonstrating the benefits of having more controlled convergence guarantees. 

One may note as well that the combination of online learning and ELM was not responsible for lower errors, showing that adding more complexity to the problem does not always translate to better solutions, possibly because our recursive proposal is already flexible enough.

\begin{table}[!ht]
    \caption{Performance Metrics for Experiment I \label{tab:table1}}
    \centering
    \begin{tabular}{l l l}
    \hline
        {Learning Method} & {RELRMSE} & {RELMAE} \\ \hline
        \textbf{MT-OSLSSVR} & \textbf{0.7498} & \textbf{0.7308} \\ 
        \textbf{MT-WRLS} & \textbf{0.7502} & \textbf{0.7310} \\
        \textbf{MT-WRLS + ELM} & \textbf{0.7700} & \textbf{0.7550} \\ 
        WRLS & 0.8467 & 0.8236 \\ 
        WRLS + ELM & 0.8490 & 0.8329 \\  
        OSLSSVR & 0.8892 & 0.8214 \\ 
        MOGD + ELM & 0.8981 & 0.8780 \\
        MOGD & 0.9336 & 0.9191 \\ 
        MADMM & 1.1244 & 1.0105 \\ \hline
    \end{tabular}
\end{table}

In Table \ref{tab:table2}, we display the outcome of our hypothesis tests for Experiment I, corroborating to the results presented before. It shows that our online MTL proposals are responsible for the highest amount of victories with the lowest mean ranks, implying in statistical significance of their superiority. 

We can not tell, however, which of our two proposals performed better as there was not sufficient statistical evidence to reject the hypothesis of their mean ranks being equable.

\begin{table}[!ht]
    \caption{Results of Hypothesis for Experiment I \label{tab:table2}}
    \centering
    \begin{tabular}{l l l l}
    \hline
    Learning Method & \# Victories & \# Defeats & Mean Rank \\ \hline
    \textbf{MT-WRLS} & \textbf{6} & \textbf{0} & \textbf{1.77} \\ 
    \textbf{MT-OSLSSVR} & \textbf{6} & \textbf{0} & \textbf{2.00} \\ 
    \textbf{MT-WRLS + ELM} & \textbf{6} & \textbf{0} & \textbf{2.70} \\ 
    WRLS & 2 & 3 & 5.13 \\ 
    WRLS + ELM & 2 & 3 & 5.40 \\ 
    OSLSSVR & 1 & 3 & 6.10 \\ 
    MOGD + ELM & 1 & 3 & 6.37 \\ 
    MOGD & 0 & 5 & 7.37 \\ 
    MADMM & 0 & 7 & 8.17 \\ 

 \hline
    \end{tabular}
\end{table}

Tables \ref{tab:table3} and \ref{tab:table4} show the mean metrics and the outcome of hypothesis tests for Experiment 2. We draw similar analysis, endorsing the superiority of our proposals in comparison to online STL and other online MTL contenders. 

We verify the relatedness of our proposals by their similar mean RELRMSE and mean RELMAE, evidencing, in practice, that both methods MT-WRLS and MT-OSLSSVR resort to the same MTL formulation. 

\begin{table}[!ht]
    \caption{Performance Metrics for Experiment II \label{tab:table3}}
    \centering
    \begin{tabular}{l l l}
    \hline
    Learning Method & RELRMSE & RELMAE \\ \hline
    \textbf{MT-WRLS} & \textbf{0.7470} & \textbf{0.7274} \\ 
    \textbf{MT-OSLSSVR} & \textbf{0.7476} & \textbf{0.7279} \\ 
    \textbf{MT-WRLS + ELM} & \textbf{0.7600} & \textbf{0.7406} \\ 
    WRLS + ELM & 0.8336 & 0.8196 \\ 
    WRLS & 0.8406 & 0.8195 \\
    OSLSSVR & 0.8765 & 0.8170 \\ 
    MOGD + ELM & 0.9023 & 0.8824 \\ 
    MOGD & 0.9320 & 0.9180 \\ 
    MADMM & 1.1704 & 0.9779 \\ \hline

    \end{tabular}
\end{table}

\begin{table}[!ht]
    \caption{Results of Hypothesis Test for Experiment II \label{tab:table4}}
    \centering
    \begin{tabular}{l l l l}
    \hline
    Learning Method & \# Victories & \# Defeats & Mean Rank \\ \hline
    \textbf{MT-WRLS} & \textbf{6} & \textbf{0} & \textbf{1.8700} \\ 
    \textbf{MT-OSLSSVR} & \textbf{6} & \textbf{0} & \textbf{2.2300} \\ 
    \textbf{MT-WRLS + ELM} & \textbf{6} & \textbf{0} & \textbf{2.5000 }\\ 
    WRLS & 2 & 3 & 5.3300 \\ 
    WRLS + ELM & 2 & 3 & 5.1000 \\ 
    OSLSSVR & 1 & 3 & 5.9300 \\ 
    MOGD + ELM & 1 & 3 & 6.2700 \\
    MOGD & 0 & 5 & 7.2700 \\ 
    MADMM & 0 & 7 & 8.5000 \\ 
 \hline
    \end{tabular}
\end{table}

\begin{figure*}[!t]
\centering
\includegraphics[width=6in]{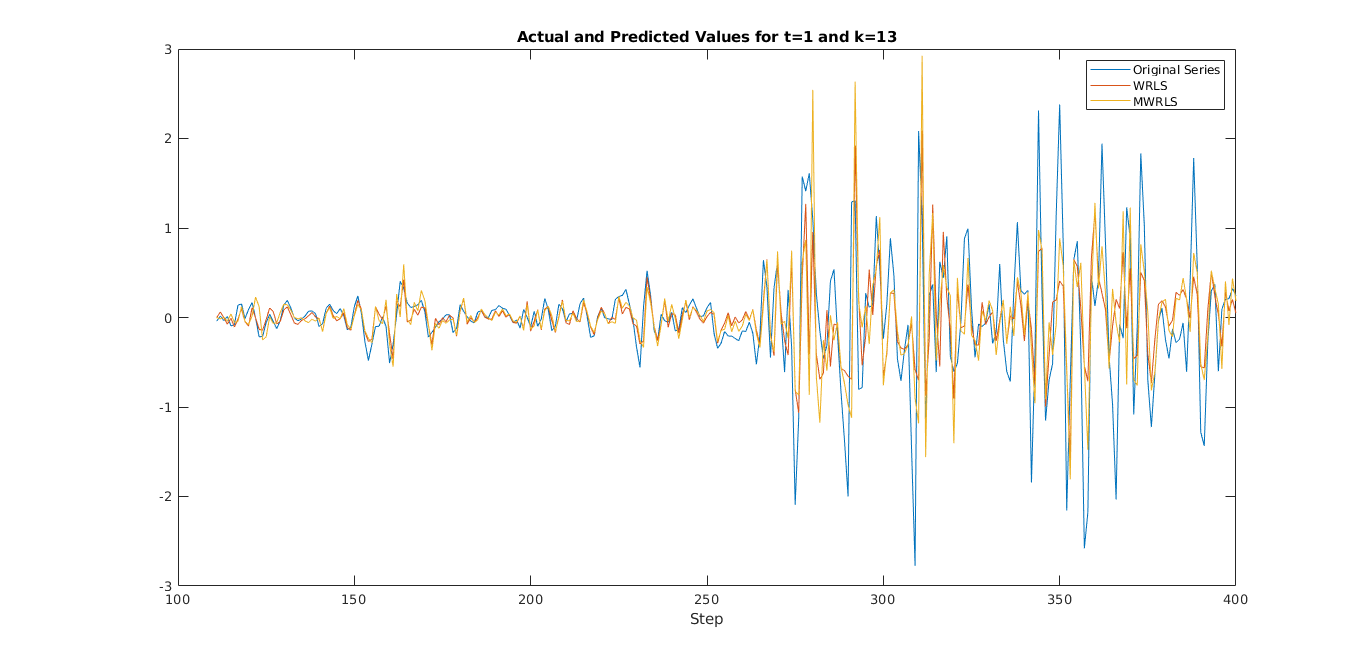}
\caption{Actual and Predicted values for time series $1$ and subset $C_{13}$ of Experiment I.}
\label{fig3}
\end{figure*}

\begin{figure*}[!t]
\centering
\includegraphics[width=6in]{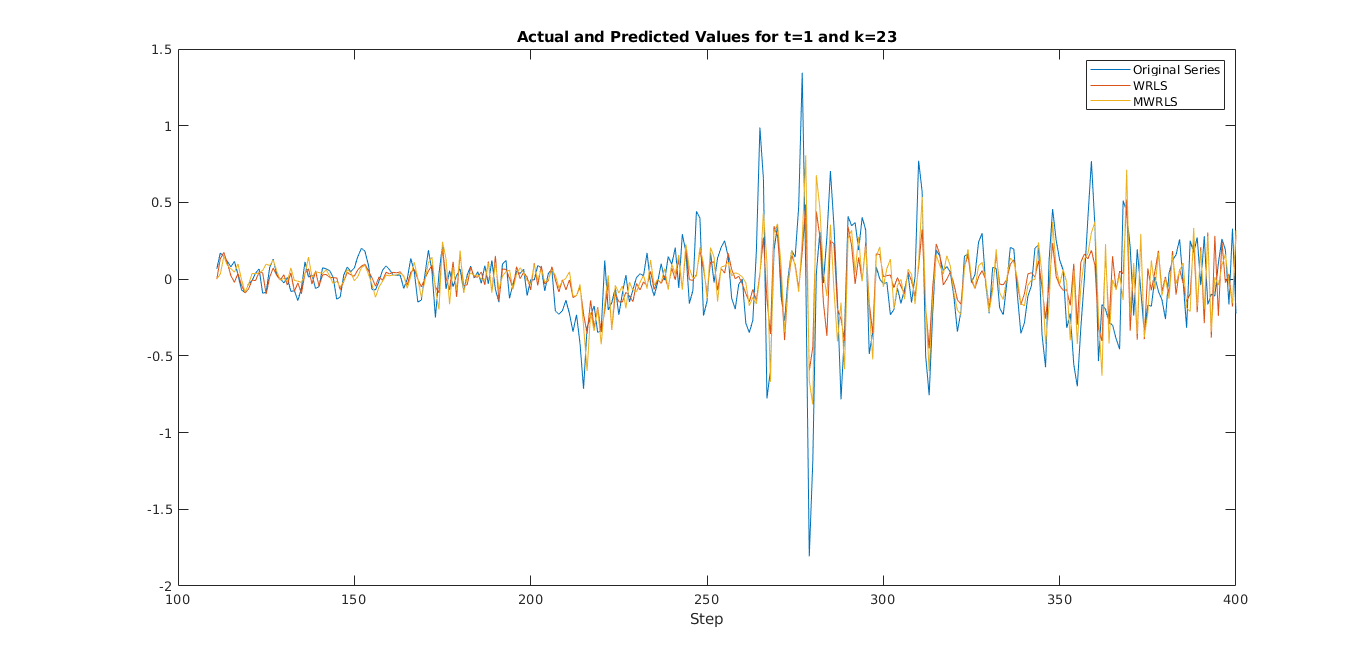}
\caption{{Actual and Predicted values for time series $1$ and subset $C_{23}$ of Experiment I.}}
\label{fig4}
\end{figure*}

Fig. \ref{fig3} and \ref{fig4} illustrate the benefits of one of our proposals, the MT-WRLS method, against its single-task version for time series (task) $1$ and subsets $C_{13}$ and $C_{23}$ of Experiment I. The more appealing gains are located in the regions of higher frequencies and higher variations, where the online MTL method was able to provide more accurate predictions.

\section{Conclusion and Future Steps}

We successfully developed two online MTL methods, one based on recursive least squares (MT-WRLS) and the other based on recursive kernel methods (MT-OSLSSVR). These two methods were proposed together due to their equivalence to the batch-fashioned multi-task graph-based formulation, as we properly deduce that the MT-WRLS and the MT-OSLSSVR provide online solutions to the same optimization problem, either in the primal or in the dual space, respectively. Profiting from efficient recursive procedures, they achieve an optimal solution at each iteration, in the case of MT-WRLS, or approximate it with controllable precision through the sparsity parameter $v$ in the case of MT-OSLSSVR. 

The per-instance cost of our proposals ($O(d^2\times T^2)$ and $O(d \times m_{n}^{2}$)) are competitive when compared to other methods in the literature, which are either based on OGD, with lower cost $O(d\times T)$ and lower convergence, or on more complex optimization procedures ($O(d^3\times T)$), whose convergence is guaranteed but not immediate such as the MADMM. Therefore, our proposals present a superior compromise between computational complexity and convergence. 

We experimented the MT-WRLS and the MT-OSLSSVR on a real-world time-dependent regression benchmark of wind speed forecasting and evidenced the higher-ranking performance of our proposals against online STL and MTL contenders, with up to $16\%$ of error reduction. Our results were submitted to statistical tests, providing statistical significance to the fact that our proposals were consistently better ranked.  

The combination of online methods with Extreme Learning Machines was also proposed to experiment the potential of nonlinear online MTL methods, a novelty in the literature of online MTL, with positive gains depending on the association.  This topic shall be better explored in the future steps of this research, specially with nonlinear multi-task kernels.

\bibliographystyle{IEEEtranN}
\bibliography{IEEEabrv,sample}

\section{Biography Section}
\vspace{11pt}
\vspace{-33pt}
\begin{IEEEbiography}[{\includegraphics[width=1in,height=1.25in,clip,keepaspectratio]{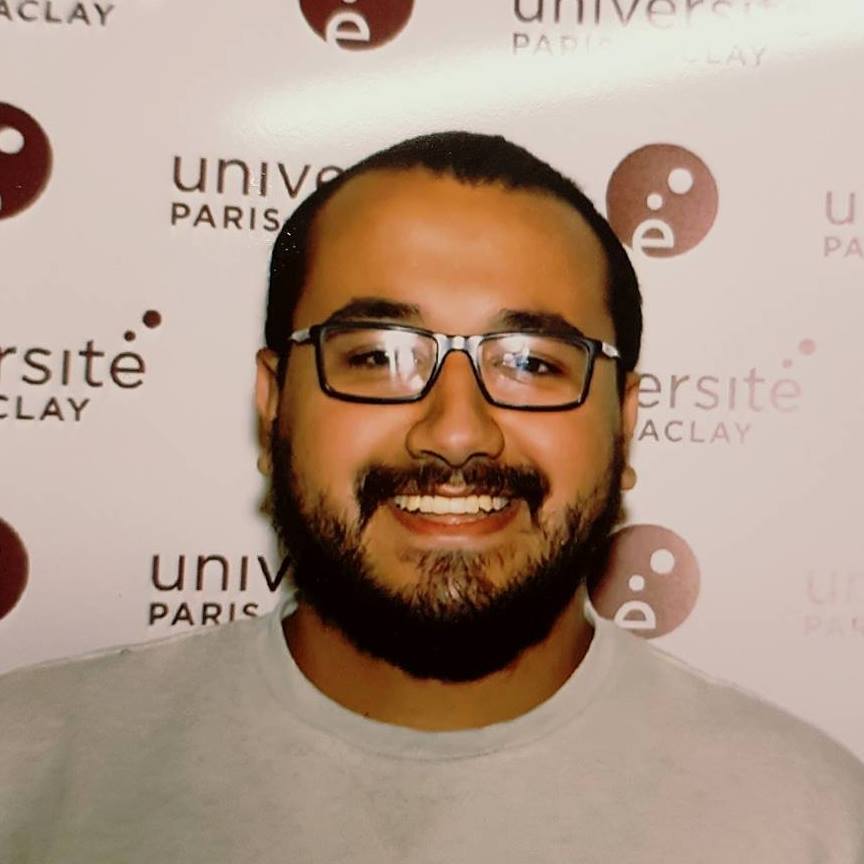}}]{Gabriel R. Lencione}
   has received his B.S. degree in Computer Engineering from the School of Electrical and Computer Engineering, University of Campinas (Unicamp), Brazil, in 2021. He also holds an M.E. degree in General Engineering from CentraleSupelec, France, obtained in 2021.
  He is currently pursuing the M.Sc. degree in Electrical Engineering with the School of Electrical and Computer Engineering, Unicamp.
  His main research interests are machine learning, computational intelligence, online learning and multi-task learning.
\end{IEEEbiography}

\vspace{-22pt}

\begin{IEEEbiography}[{\includegraphics[width=1in,height=1.25in,clip,keepaspectratio]{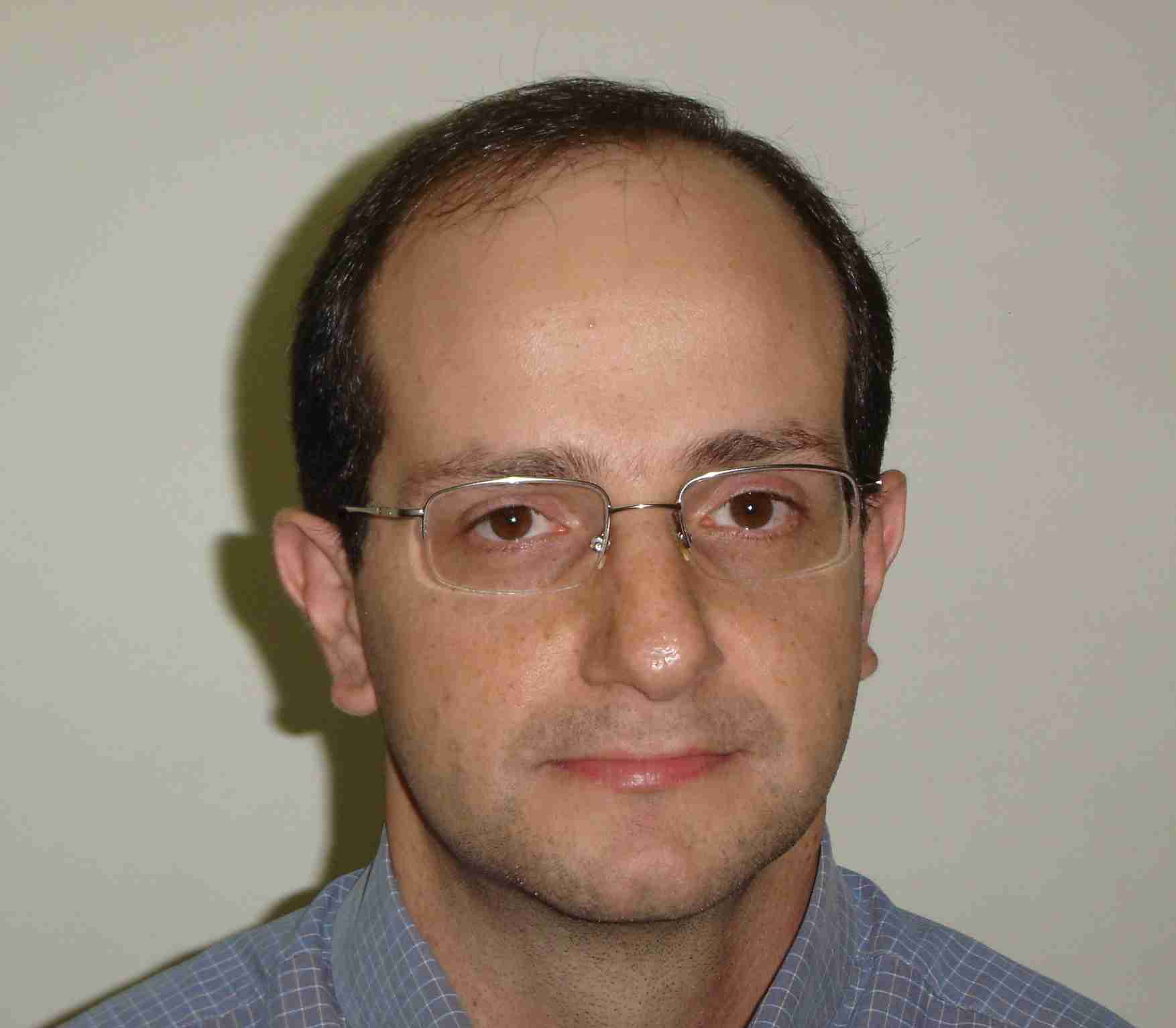}}]{Fernando J. Von Zuben}
   is a Full Professor at the Department of Computer Engineering and Industrial Automation, School of Electrical and Computer Engineering, University of Campinas (Unicamp), São Paulo, Brazil. The main topics of his research are computational intelligence, bioinspired computing, multivariate data analysis, and machine learning. He coordinates open-ended research projects on these topics, tackling real-world problems in information technology, decision-making, pattern recognition, and discrete and continuous optimization.
\end{IEEEbiography}

\vfill

\end{document}